# Virtual Sensor Modelling using Neural Networks with Coefficient-based Adaptive Weights and Biases Search Algorithm for Diesel Engines

Kushagra Rastogi
Department of Electrical and Computer Engineering
University of California, Los Angeles
Los Angeles, United States
kushagra02rastogi@yahoo.co.in

Navreet Saini
Electronics Engineer, Cummins EBU-Controls
Cummins India Ltd
Pune, India
navreet.saini@cummins.com

*Abstract* — With the explosion in the field of Big Data and introduction of more stringent emission norms every three to five years, automotive companies must not only continue to enhance the fuel economy ratings of their products, but also provide valued services to their customers such as delivering engine performance and health reports at regular intervals. A reasonable solution to both issues is installing a variety of sensors on the engine. Sensor data can be used to develop fuel economy features and will directly indicate engine performance. However, mounting a plethora of sensors is impractical in a very cost-sensitive industry. Thus, virtual sensors can replace physical sensors by reducing cost while capturing essential engine data.

## I. INTRODUCTION

Advanced technology resides in automobiles to comply with the stringent emission norms. This is achieved using an electronic control module (ECM) which obtains data from sensors present in the engine and vehicle. However, a lot of sensors need to be present on the engine to closely monitor its health and performance. This has several drawbacks. In a price-sensitive industry like automotive, adding sensors to the engine increases the cost the product. This repels customers immediately. Another major problem is fitment constraints. It may not be possible to fit sensors on certain sections of the engine or it may require modifications in the structure of the engine to fit the sensor. This disrupts the design process and creates hassle.

A suitable alternative to a physical sensor is a virtual sensor. It has several advantages over its hardware counterpart. The biggest advantage is the presence of a virtual sensor eradicates the need for hardware; this reduces cost without substantially compromising the quantity and quality of the collected data. Implementation of virtual sensors also allows for a simpler design of the engine and ECM; the engine will not be overloaded with extraneous physical sensors. Lastly, virtual sensors are software-oriented and data-driven models; this makes them much easier to control and change.

Therefore, virtual sensors are more flexible and promote the increasingly popular phenomena of Big Data and data analytics.

## II. VIRTUAL SENSOR MODELLING

Virtual sensors are mathematical models to approximate the behavior of a physical sensor [1]. For instance, assume there are four sensors x1, x2, x3, and x4. If x4 needs to be converted into a virtual sensor, it can be mathematically represented as

$$x4 = f(x1, x2, x3) \qquad (1)$$

It should be noted that x4 could potentially be a function of all the other sensors, some of the other sensors or none of the other sensors. A correlation (linear or non-linear) must exist between the inputs and the virtual sensor. If no correlation exists, then creating a virtual sensor is not possible. Hence, virtual sensor modelling comprises of two main processes: identifying relevant inputs and developing the modelling technique/algorithm. This paper will not extensively discuss how to identify relevant inputs to the virtual sensor. However, a general step-wise outline is given below:

1. Classify the virtual sensor as a section in context of the main system being analyzed. Example: Oil pressure is related to combustion chamber.

2. Qualitatively describe the relationship between virtual sensor and surrounding parameter. Example: Oil pressure increases as engine speed increases.

3. Use specialized knowledge and consult appropriate teams/people to eliminate potentially irrelevant inputs

4. Perform statistical analysis

5. Repeat steps 3 & 4 to until a set of **independent** set of relevant inputs are chosen

Once a good set of appropriate inputs are selected, the modelling technique must be decided. Traditionally for function approximation tasks, techniques such as linear regression, weighted least squares etc. have been employed. These techniques, by their definitions, are constructed on the axioms of linearity and in many cases, non-linear relationships can be estimated using these linear models. But this puts dents in the quality of the prediction in terms of accuracy and reliability. It is difficult to achieve good results when linear models are applied to approximate highly volatile, non-linear relationships.

If the virtual sensor is a data-driven model instead of a physics-based model, it can be hard to exactly pinpoint the relationship between the inputs and the virtual sensor before and during the modelling process. Hence, using linear techniques may not be apt when the relationship between the inputs and virtual sensor is unknown. As a result, the modelling technique must be robust in the sense that it must be able to sufficiently handle the linearities and non-linearities in the relationship between inputs and virtual sensor. Neural networks have this desired capability and can be exploited for function approximation purposes.

### III. Neural network structure

A neural network is a network of interconnected information processing units (neurons) that can be programmed to do specific tasks like function approximation, image classification and speech recognition [2].

A neural network solution is devised to perform virtual sensor modelling. The solution is developed in MATLAB. The solution consists of two neural networks: two feedforward neural networks with different training algorithms. The topology was chosen based on its relative success at regression tasks.

#### A. Feedforward Networks

Two feedforward networks are created. The only differences between them are the number of neurons, weights and bias assignments and training algorithm. One feedforward network uses the Levenberg-Marquardt backpropagation training algorithm (hereafter referred to as 'trainlm') and the other network implements the Bayesian regularization backpropagation training algorithm (hereafter referred to as 'trainbr'). These training algorithms were selected on speed, accuracy and memory considerations. The following analysis and discussion are common to both feedforward networks and hence they will be referred to as 'the network'.

The network has three layers: an input layer, one hidden layer and an output layer. The input layer consists of a concise number of relevant inputs that are crucial for the virtual sensor. The output layer contains one neuron since only virtual sensor will be modelled at any instant of time. Lastly, one hidden layer was chosen because it can adequately deal with linearities and most non-linearities in the virtual sensor-inputs relationship. The number of neurons in the hidden layer is calculated through a complex decision-making process.

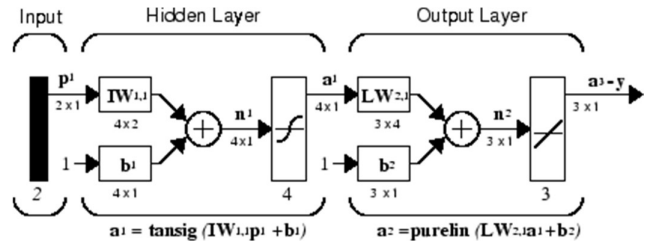

Fig 2: Diagram showing the architecture of the network including the activation functions for the hidden and output layers [3]

The input and target datasets for the network are divided into training, validation and testing set using specific ratios: training ratio = 60 %, validation ratio = 20 %, testing ratio = 20 %. The data division function used is interleaved division or 'divideint' because it trains the network on the entire range of the input dataset which results in better generalization.

The initial weights and biases are configured at the start. The following quantities are hardcoded: input weights (IW), weights of hidden layer (LW), bias of hidden layer (B1) and bias of output layer (B2). These quantities are labelled in Fig 2. Manually entering the weights and biases is acceptable because all four quantities are subject to change after backpropagation is executed. The network is tested on six different set of configured weights and biases. They are –

Set 1: IW = 1, B1 = 1, B2 = 1, LW = 0
Set 2: IW = 1, B1 = 1, B2 = 1, LW = 1
Set 3: IW = 1, B1 = 0, B2 = 0, LW = 1
Set 4: IW = 1, B1 = 1, B2 = 0, LW = 1
Set 5: IW = 1, B1 = 0, B2 = 1, LW = 1
Set 6: IW = 0, B1 = 1, B2 = 1, LW = 1

Every configured weight and bias is a matrix of the proper dimensions.

The number of neurons for each set are run in a loop from 2 to 50 with a step size of 1. The optimal number of neurons is determined for each set of configured weights and biases using two measures:

1. perf = Mean squared error between target data and predicted virtual sensor output

2. countPercent = Percentage of data points in predicted virtual sensor output that have an accuracy equal to or greater than 99 %.

Each set of configured weights and biases has these two measures. The process of deciding the number of neurons for the set is as follows: if the difference between the maximum countPercent and the countPercent corresponding to the minimum perf is greater than 2, neurons corresponding to the maximum countPercent are chosen. Otherwise, neurons corresponding to the minimum perf are chosen. This procedure is completed for each set of configured weights and biases. In the end, an optimal number of neurons and its corresponding perf and countPercent is obtained for each set of configured weights and biases. Therefore, three arrays can be formed: an array containing six entries denoting number of neurons, an array for the six corresponding perf values and an array for the six corresponding countPercent values.

Once the array containing the optimal number of neurons of each set is determined, the problem shifts to finding the best number of neurons within that array. The criteria for choosing the number of neurons are –

i) Choose the smallest number of neurons

ii) Choose the number of neurons with the minimum perf

iii) Choose the number of neurons with the maximum countPercent

All three quantities (neurons, perf and countPercent) are trying to be optimized with the constraints that number of neurons and perf should be minimized whereas countPercent should be maximized. As a result, this is almost analogous to a constrained optimization problem in three variables. It is tough to solve. Hence, three thresholds were developed to aid in the decision-making. They are –

1) neuronCut = 5

2) perfCut

3) countCut

The value of perfCut is –

$$\bar{x}\,[std\,(pA), \max\,(pA) + \min\,(pA) - 2 \times (\overline{pA})\,]\quad(2)$$

The value of countCut is –

$$\bar{x}\,[std\,(cA), \max\,(cA) + \min\,(cA) - 2 \times (\overline{cA})\,]\quad(3)$$

pA represents the array containing all the 6 perf values. cA represents the array containing all the 6 countPercent values.

All the six configured weights and biases were compared against each other using these thresholds and a final decision on the best number of neurons was made.

After the number of neurons is finalized, the network is run on the input and the target datasets. Two additional parameters are monitored to judge the performance of the network.

3. range = difference in accuracy between most accurate data point and least accurate data point in the predicted virtual sensor output

4. R-sq = regression coefficient signifying the relationship between target data and predicted virtual sensor output

The initial results for both feedforward networks are presented below for an Oil Pressure virtual sensor.

Feedforward network with 'trainlm' training algorithm:

a) Neurons = 38

b) Perf = 1.0715 kPa

c) Range = 0.9290 %

d) countPercent = 100 %

e) R-sq = 0.9999

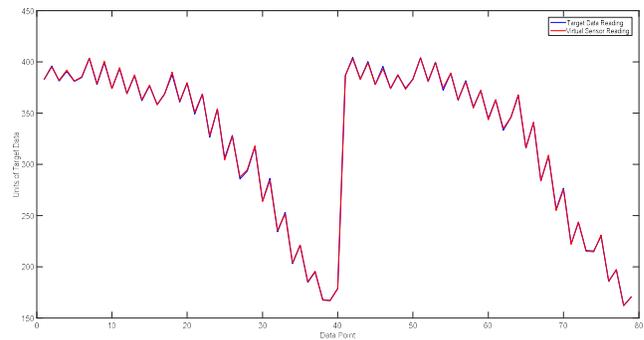

Fig 3: Initial plot of target data (blue) vs predicted virtual sensor output (red) of feedforward network with 'trainlm' for Oil Pressure

Feedforward network with 'trainbr' training algorithm:

a) Neurons = 28

b) Perf = 0.6224 kPa

c) Range = 0.7080 %

d) countPercent = 100 %

e) R-sq = 0.9999

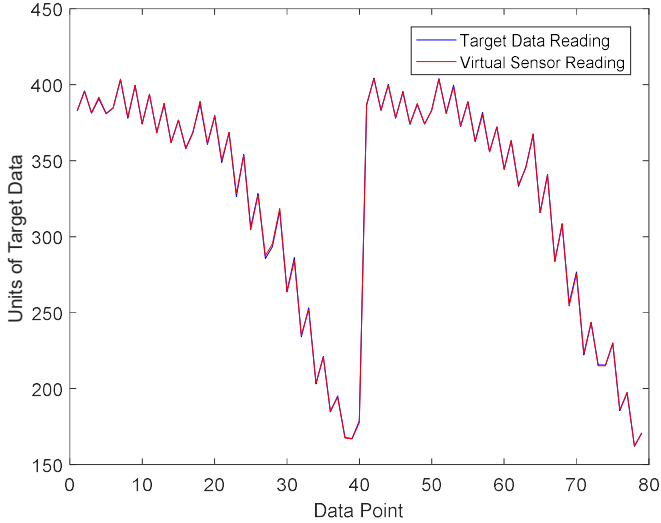

Fig 4: Initial plot of target data (blue) vs predicted virtual sensor output (red) of feedforward network with 'trainbr' for Oil Pressure

As it can be seen from the statistics shown above, the initial results for both networks look quite promising. This can be corroborated with Fig 3 and Fig 4 which depict the fact that the predicted virtual sensor output almost largely overlaps the target data. This suggests that the networks have small generalization errors.

The initial results for the networks look good. However, feedforward have a tendency to get stuck in local optimum instead of reaching global optimum. Therefore, the performance of the networks can be improved by configuring the initial weights and biases to provide better initial conditions for training and generalization. The AWB algorithm (discussed in the next main section) is used to modify the originally configured weights and biases.

## IV. AWB ALGORITHM

The algorithm is a simple coefficient-based adaptive algorithm built on the principles of numerical approximation and inspired by the concept of the divide-and-conquer [4]. It is similar to grid search hyperparameter optimization techniques. It is used to tune the manually configured weights and biases of the feedforward networks.

The input to the algorithm is the coefficient of a weight/bias matrix and the original performance parameters. The output of the algorithm is an adjusted coefficient which guarantees an improvement in performance-judging parameters by 0 % or greater.

The algorithm starts with a fixed search space and then iteratively reduces it by using performance parameter indices to direct the search. There are three iterations. The algorithm does not majorly consider its computational time and complexity because it is developed solely for improving accuracy and performance.

### A. First Iteration

On the first iteration, the search space is fixed from -5 to 5 with step size 0.1 for the selected quantity (IW, B1, B2 or LW). A larger search space was not selected because it is not recommended to use large weights in the network. A feedforward network is then run with the selected quantity ranging from -5 to 5 with step size 0.1. Four performance parameters (perf, counPercent, range and R-sq) are monitored to understand which coefficient produces the best result. The tendency of the algorithm is to choose the coefficient corresponding to index where perf is minimized. However, if perf index does not provide the best results, then the picking order becomes

1. Index where range is minimized
2. Index where countPercent is maximized
3. Index where R-sq is maximized

Once the index is decided, the coefficient corresponding to that index is determined. Following this, the algorithm moves onto the second iteration.

### B. Second Iteration

In the second iteration, the value of the coefficient from the first iteration is very important because it elects the size of the search space. If the coefficient is between 0 and 2.5 or 0 and -2.5, then the search space becomes 0 to 2.5 or 0 to -2.5 respectively. Otherwise, the search space becomes 2.5 to 5 or -2.5 to -5. The step size in both cases is 0.01. After the search space has been reduced through the value of the coefficient, the network is run again, and the four performance parameters are checked to know which index to choose. If the performance at this iteration is worse than the performance of the first iteration, then the search space changes to, in general, (-1*coefficient + 0.1) to (coefficient + 0.1) or (coefficient – 0.1) to (-1*coefficient – 0.1) depending on the sign of the coefficient. The step size remains 0.01. The search space is slightly different for boundary conditions (coefficient is equal to -5, 0 or 5).

The change in the search space is caused by the adaptive feature of the algorithm. The algorithm realizes that it is not travelling along the best path and thus it searches for a better path. Furthermore, the algorithm will never go down the wrong path again (unless the wrong path is the best path available) because it learns from past experiences and recognizes a wrong path when it encounters one. After the search space is adaptively modified, the network is run and the best index is selected. The coefficient corresponding to this index is determined and the algorithm moves to the third iteration.

*C. Third Iteration*

The third iteration is the last iteration and its function is to aggressively narrow-down the coefficient. In the third iteration, the value of the coefficient from the second iteration is used to decide the search space. The search space becomes, in general, (coefficient – 0.01) to (coefficient + 0.01). The step size changes to 0.0001. The network is run and the best index is selected. The coefficient corresponding to this index is temporarily labelled as the final, adjusted and adapted coefficient. The algorithm stops after the third iteration and verifies if the final, adapted performance parameters are satisfactorily better than the original performance parameters. If the adapted parameters are better than the original ones, then the calculated coefficient is declared as the final, adapted coefficient. Otherwise, the original coefficient remains unchanged.

After the coefficient of one quantity has been adapted, the coefficients of the other quantities are also adapted. In this way, the performance and accuracy of the network is potentially maximized.

*D. Minor Change in Algorithm*

The inner workings of the algorithm described above apply and hold true for both feedforward networks. However, there is one change in the algorithm for the feedforward network with training algorithm 'trainbr'. 'trainbr' is slower than 'trainlm'. To counteract and compensate for this, the step sizes are smaller. For the feedforward network with 'trainbr', the following step sizes are used: first iteration = 0.5, second iteration = 0.05 and third iteration = 0.005.

*E. Final Results*

Final results for feedforward network with 'trainlm', after the AWB algorithm is run, are displayed below for an Oil Pressure virtual sensor.

a) Perf = 0.8958 kPa

b) Range = 0.7116 %

c) countPercent = 100 %

d) R-sq = 0.9999

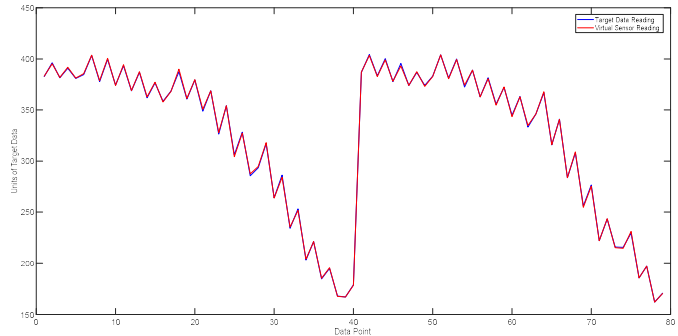

Fig 7: Final plot of target data (blue) vs predicted virtual sensor output (red) of feedforward network with 'trainlm' for Oil Pressure

Final results for feedforward network with 'trainlm', after the AWB algorithm is run, are displayed below for an Oil Pressure virtual sensor.

a) Perf = 0.5961 kPa

b) Range = 0.7007 %

c) countPercent = 100 %

d) R-sq = 0.9999

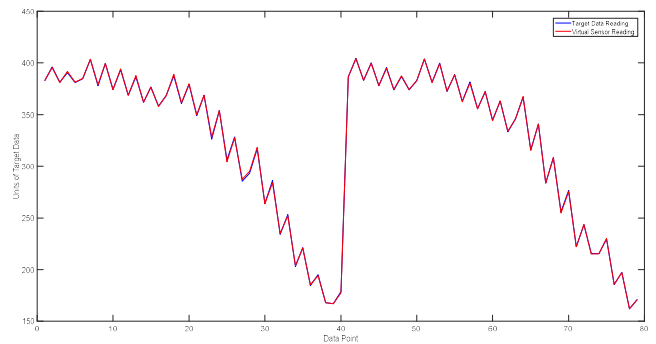

Fig 8: Final plot of target data (blue) vs predicted virtual sensor output (red) of feedforward network with 'trainbr' for Oil Pressure

If the above results are compared with the initial results for both networks, it can clearly be observed that the accuracy for both networks has ameliorated. Most notably, the range of the feedforward network with 'trainlm' has decreased by 23 %. This signifies that the predicted virtual sensor output data points are closer together and closer to 100 % accuracy.

V. CONCLUSION

A neural network solution, consisting of three networks, was proposed to accomplish the task of virtual sensor modelling. The predicted output of all three networks was quite accurate. An algorithm was developed to tune the manually configured weights and biases of the feedforward networks. The algorithm helped improve the accuracy of the networks, in one case by as much as 23 %. Thus, the proposed neural network is effective in carrying out its purpose. Overall, neural networks are more robust modelling tools than linear regression equations because neural networks are dynamic whereas linear regression equations are static. Virtual sensors in highly dynamic environments like a diesel engine should be and can only be modelled to their fullest extent using their dynamic modelling equivalent, neural networks.

ACKNOWLEDGEMENT

We would like to thank Cummins India Ltd for providing and allowing access to its resources.